\pdfoutput=1

\documentclass[11pt]{article}

\usepackage[]{acl}

\usepackage{times}
\usepackage{latexsym}
\usepackage{enumitem}
\usepackage{url}

\usepackage[T1]{fontenc}

\usepackage[utf8]{inputenc}

\usepackage{microtype}

\usepackage{inconsolata}

\usepackage{xcolor}
\usepackage{tikz-dependency}
\usepackage{booktabs}
\usepackage[export]{adjustbox}
\usepackage{multirow}
\usepackage{bbding}
\usepackage{amsmath}

\usepackage{tabularx}

\usepackage{listings}
\usepackage{float}
\usepackage{subcaption}

\title{Label-semantics Aware Generative Approach for Domain-Agnostic Multilabel Classification}

\author{
    Subhendu Khatuya,
    Shashwat Naidu,
    Saptarshi Ghosh,
    Pawan Goyal,
    Niloy Ganguly \\
Indian Institute of Technology Kharagpur, India\\
    subha.cse143@gmail.com, shashwatnaidu07@gmail.com, \\ saptarshi@cse.iitkgp.ac.in, pawang@cse.iitkgp.ac.in, niloy@cse.iitkgp.ac.in
}

\newcommand{\noteng}[1]{\textcolor{red}{\bf\small [#1 --NG]}}

\newcommand{\notesk}[1]{\textcolor{blue}{\bf\small [#1 --SK]}}

\newcommand{\modelname}{LAGAMC}

\begin{document}
\maketitle
\begin{abstract}

 The explosion of textual data has made manual document classification increasingly challenging. To address this, we introduce a robust, efficient domain-agnostic generative model framework for multi-label text classification. Instead of treating labels as mere atomic symbols, our approach utilizes predefined label descriptions and is trained to generate these descriptions based on the input text. During inference, the generated descriptions are matched to the predefined labels using a finetuned sentence transformer. We integrate this with a dual-objective loss function, combining cross-entropy loss and cosine similarity of the generated sentences with the predefined target descriptions, ensuring both semantic alignment and accuracy.
Our proposed model \modelname~stands out for its parameter efficiency and versatility across diverse datasets, making it well-suited for practical applications. We demonstrate the effectiveness of our proposed model by achieving new state-of-the-art performances across all evaluated datasets, surpassing several strong baselines. We achieve improvements of $\textbf{13.94}\textbf{\%}$ in Micro-F1 and $\textbf{24.85}\textbf{\%}$ in Macro-F1 compared to the closest baseline across all datasets.


\end{abstract}

\section{Introduction}
Text classification automates the analysis and organization of large datasets, enabling efficient labeling, categorization, and valuable insights. In text classification, two main categories arise: multi-class and multi-label classification. Multi-class classification assigns a single category to a text, while multi-label classification (MLTC) \cite{xiao2019label}  assigns multiple relevant labels to a document. Real-world applications include topic recognition \cite{yang2016hierarchical}, question answering \cite{kumar2016ask}, sentiment analysis \cite{senticnet}, information retrieval \cite{Gopal2010MultilabelCW}, and text categorization \cite{boosTexter}, among others. 



In the field of multi-label text classification, various approaches including traditional machine learning, deep learning, and hybrid models have been proposed \cite{chen2022survey}. Several state-of-the-art methods \cite{huang2021balancing, xiao2019label} emphasize the importance of label descriptions or metadata in improving model performance and capturing label correlations. However, these approaches often face limitations in their generalizability and adaptability. Some models rely on label hierarchies or meta-path graphs \cite{ye2024matchxmlefficienttextlabelmatching}, which, although effective in certain contexts, hinder scalability and flexibility. Additionally, some models are designed for specific tasks, such as legal or financial applications, integrating label descriptions for specialized classifiers \cite{chalkidis-etal-2020-legal, khatuya-etal-2024-parameter}. Despite their advancements, these methods remain limited to particular domains, highlighting the need for a more generalizable approach to MLTC.

Large Language Models (LLMs), with their extensive pretraining, are capable of understanding similarities and relationships based on textual patterns \cite{naveed2024comprehensiveoverviewlargelanguage}. This motivated us to propose a novel domain-agnostic pipeline that leverages recent generative models in a parameter-efficient setup employing Low-Rank Adaptation (LoRA) \cite{hu2021lora}, for the MLTC task. Unlike previous approaches that treat label descriptions as metadata, our proposed approach \modelname~ trains the generative model to generate these descriptions directly. This enables us to harness the full potential of LLMs, providing a more nuanced representation of label relationships and their connections to the corresponding text.

Obtaining label descriptions manually for multi-label datasets is both labor-intensive and subjective. To overcome this, we automate the process by leveraging GPT-3.5 \cite{brown2020language} and Wikipedia, enabling the efficient creation of label descriptions with minimal human intervention.
By leveraging a generative model while providing semantic information about labels, our method consistently outperforms state-of-the-art models, achieving a \textbf{9.32\%} improvement in Micro-F1 and \textbf{15.25\%} in Macro-F1 over the closest baseline.
 

We further integrate a dual-objective loss function, combining cross-entropy loss with a cosine similarity-based loss. This results in an additional increase of \textbf{4.62\%} and \textbf{9.60\%} in Micro-F1 and Macro-F1 respectively. Using such a hybrid loss helps bring the embeddings of the outputs of the generative model and that of the predefined label description closer in the representation space, making it easier to map the generated outputs with final labels. This hybrid approach ensures the model comprehends label distinctions and avoids overfitting to token-level matches, contributing to improved performance across diverse datasets.


Empirical evaluations on diverse datasets spanning social media, news, academic, and healthcare domains demonstrate the effectiveness of our approach \modelname. Across various datasets, our method consistently outperforms state-of-the-art models, achieving an overall improvement of \textbf{13.94\%} in Micro-F1 and \textbf{24.85\%} in Macro-F1 over the closest baseline. Additionally, our model shows strong performance on rare labels and exhibits zero-shot capability, further enhancing its applicability.





\noindent In summary, the key contributions of this paper are as follows:
\begin{itemize}[leftmargin=*]
    \item We generated label descriptions for datasets lacking them, using an annotation process guided by GPT-3.5 \cite{brown2020language} and Wikipedia. Our dataset\footnote{Data: \url{https://drive.google.com/drive/folders/1nrCKgmEtYrM1mQHIu3-eKXEtRdlIUDiO?usp=sharing}} and code are publicly available at this anonymous link\footnote{Code: \url{https://github.com/subhendukhatuya/Gen_Multilabel_Classification.git}}. 
    \item We developed a novel parameter-efficient generative approach for MLTC, leveraging label descriptions to improve classification accuracy.
    \item We introduced a dual-objective loss function, incorporating a semantic similarity-based loss to enhance the model's semantic understanding.

    \item  Our method generalizes well across domains, as demonstrated by performance metrics on various datasets. For all the datasets, our proposed model \modelname, achieves huge improvements over baselines. Furthermore, our model excels on rare labels and demonstrates zero-shot capability.
\end{itemize}

\if{false}
Existing state-of-the-art models for MLTC, such as AttentionXML \cite{you2019attentionxml}, DEXA \cite{dahiya2023deep}, and DeepXML \cite{dahiya2021deepxml}, do not leverage label descriptions. These models are highly compute-intensive and tend to be limited in their ability to handle unseen labels during inference due to their discriminative approach. \fi

\if{false}
Text classification automates the analysis and organization of large datasets, enabling efficient labeling, categorization, and valuable insights.
In text classification, two main categories arise: multi-class and multi-label classification. Multi-class classification assigns a single category to a text, while multi-label classification (MLTC) \cite{xiao2019label} allows a single document to belong to multiple categories. MLTC poses a greater challenge, as it requires discerning and assigning multiple relevant labels to a document. Real-world applications include topic recognition \cite{yang2016hierarchical}, question answering \cite{kumar2016ask}, sentiment analysis \cite{senticnet}, information retrieval \cite{Gopal2010MultilabelCW}, tag recommendation \cite{Katakis2008MultilabelTC}, and text categorization \cite{boosTexter}, among others. It is essential to develop generalizable models for effectively handling diverse, real-world applications. In MLTC, semantic relationships between labels are common, as they often overlap in documents. We leverage this characteristic by incorporating label descriptions into our model. These descriptions enhance the model’s ability to understand document content and uncover correlations between similar labels, resulting in more reliable predictions \cite{huang2021balancing, ma-etal-2023-pai, xiao2019label}. 
\noteng{So this label correlation is not your contribution, somebody else have discovered it } \notesk{-- Yes we  leveraged label descriptions that into our generative frame work }

For our experiments, we selected several well-established \textit{multi-label classification datasets}. We then generated label descriptions using GPT-3.5 \cite{brown2020language} and Wikipedia.
\noteng{The dataset description can come later}
We demonstrate the utility of label descriptions through an example shown in Table \ref{tbl:examples} from the SemEval dataset \cite{mohammad2018semeval}. According to descriptions, `Sadness' is defined broadly as grief and feeling of sorrow and `Pessimism' as lack of confidence and inclination to expect negative outcome. In the first example, the absence of grief connects solely to pessimism, indicating a lack of confidence. In contrast, the second example not only reflects a lack of confidence but also anticipates negative outcomes and embodies feelings of grief. 
\noteng{I am not even sure of the example what does it tell} \notesk{Basically in Table 1 - 2nd example first few words talks about pessimism and  last few words talks about grief, now in the Sadness description there is a term grief -so model was able to tag that as both pessimism and sadness, we observed that the same example is tagged by only pessimism by other baselines. Anyway let me try to find and replace with some other example. }
These observations motivated us to integrate label descriptions into our model, significantly improving its usability, adaptability, and performance. \noteng{are you generating labels using LLMs? Entire thing is in a mess} \notesk{First we took label descriptions from wiki then modified that description using GPT-3.5 giving wiki description as initial description--check Section 3 ('anger' wiki description) and Table 3 ('anger'- final description) } We utilize Large Language Models (LLMs), pretrained on vast corpora, and along with label descriptions to build a generalizable framework for MLTC. Incorporating label descriptions enables LLMs to capture the semantic relationships between labels, enhancing contextual understanding. We further integrate a dual-objective loss function, combining cross-entropy loss with a cosine similarity-based loss. This hybrid approach ensures the model comprehends label distinctions and avoids overfitting to token-level matches, contributing to improved performance across diverse datasets. Existing state-of-the-art models for multi-label text classification (MLTC), such as AttentionXML \cite{you2019attentionxml}, DEXA~\cite{dahiya2023deep} and DeepXML \cite{dahiya2021deepxml}, do not leverage label descriptions. These models are very compute intensive and tend to be limited in their ability to handle unseen labels during inference due to their discriminative approach. While models like MATCH \cite{ye2024matchxmlefficienttextlabelmatching} incorporate metadata, they rely on label hierarchies for training, which also restricts their capacity to generalize to unseen labels at inference time. Other zero-shot MLTC models, such as MICol \cite{zhang2022metadatainducedcontrastivelearningzeroshot}, depend on constructing meta-path graphs, which are not feasible for all datasets, thereby limiting their applicability to specific domains.
In contrast, our work presents a novel domain-agnostic pipeline that leverages recent generative models in a parameter-efficient setup for the MLTC task.
While our current experiments focus on FLAN-T5, the framework's flexible architecture allows for seamless incorporation of newer LLMs in future. 

Additionally, for domain-specific MLTC needs, our pipeline allows the use of LLMs tailored to specific domains, offering flexibility and robustness for both general and specialized tasks. The proposed model also exhibits zero-shot capabilities. To reduce computational costs, we employ Low-Rank Adaptation of LLMs (LoRA~\cite{hu2021lora}), which significantly reduces the number of trainable parameters and GPU memory requirements. As a result, our approach delivers a robust and efficient solution for MLTC. Across various datasets from social media, news, academic, and healthcare domains, our method consistently outperforms state-of-the-art models, achieving an improvement of $\textbf{13.08}\textbf{\%}$ in Micro-F1 and $\textbf{24.85}\textbf{\%}$ in Macro-F1 over the closest baseline.

\noindent In summary, the key contributions of this paper are as follows:
\begin{itemize}[leftmargin=*]
    \item We generated label descriptions for datasets lacking them, using an annotation process guided by GPT-3.5 \cite{brown2020language} and Wikipedia. Our dataset\footnote{Data: \url{https://drive.google.com/drive/folders/1nrCKgmEtYrM1mQHIu3-eKXEtRdlIUDiO?usp=sharing}} and code are publicly available at this anonymous link\footnote{Code: \url{https://anonymous.4open.science/r/GenerativeMultiLabel_Classification-5415/README.md}}. 
    \item We developed a novel parameter-efficient generative approach for MLTC, leveraging label descriptions to improve classification accuracy.
    \item We introduced a dual-objective loss function, incorporating a semantic similarity-based loss to enhance the model's semantic understanding of the labels.
    \item Our method generalizes well across diverse domains, as demonstrated by performance metrics on various datasets.

     \item For all the datasets, our proposed model \modelname, achieves huge improvements over all the baselines. Additional advantages of our model include substantial performance over rare labels, and its zero-shot capability.

\end{itemize}

\fi

\if{0}
\section{Related Works}
In literature, this task has been approached by means of several different techniques, ranging from simple single-label classifier methods to neural network-based models, or through the adoption of transformer-based approaches. The approaches to solving this problem can be classified into three approaches: the transformation of the problem, algorithm adaption and neural network based models \cite{MLTC}. 
The transformation approach to the problem is algorithm-independent, that is, it transforms the multi-label learning task into a single-label classification task but this is computationally intractable and impractical for the amount of data we have in real world today. 
The second approach involves directly handling the multi-label data by modifying some particular learning models and algorithms. For example, using decision trees with multiple labels in the leaves of the tree. Various neural network-based methods such as CNN \cite{Liu2017DeepLF} \cite{kurata-etal-2016-improved}, RNN \cite{liu2016recurrent}, combination of CNN and RNN \cite{Lai_Xu_Liu_Zhao_2015} \cite{7966144}, attention mechanism \cite{yang2016hierarchical} \cite{you2019attentionxml} \cite{adhikari2019docbert} etc., have achieved great success in document representation.  All of the models mentioned above majorly focus on document representation and tend to ignore the label semantics and content. In recent times, transformer-based approaches are being used to solve this task. Pretrained transformer models like  BERT \cite{devlin2019bert}, RoBERTa \cite{liu2019roberta} are being utilized to build a neural network-based approach for multi-label text classification. 
In \cite{AMEER2023118534}, authors propose building multiple attention layers over the final layer of RoBERTa to make the final prediction. Some methods including DXML \cite{zhang2018deep}, EXAM \cite{du2018explicit}, SGM \cite{yang2018sgm}, GILE \cite{pappas2019gile} are proposed to capture the label correlations by exploiting label structure or label content. 





\fi


\section{Related Works}
Multi-label text classification has been approached through a range of techniques, including extending single-label classifiers, employing neural network architectures, and, more recently, utilizing transformer and pretrained language models based works. Neural network-based approaches have shown great success, with methods leveraging CNNs \cite{Liu2017DeepLF}, RNNs \cite{liu2016recurrent}, and hybrid CNN-RNN models \cite{7966144}. Attention-based models \cite{yang2016hierarchical, you2019attentionxml, adhikari2019docbert} have also improved document representation, but often overlook label semantics and dependencies. 

More recent approaches utilize pretrained transformers such as BERT \cite{devlin2019bert} and RoBERTa \cite{liu2019roberta}, which have been adapted for multi-label tasks. For instance, \cite{AMEER2023118534} applies multiple attention layers over RoBERTa’s final layer to enhance label correlation learning. \citep{ma-etal-2023-pai} explores various loss functions designed to mitigate the impact of class imbalance in datasets.

Most of these do not leverage the information contained in label descriptions. 
Some of the works that utilize label descriptions to enhance performance \cite{ye2024matchxmlefficienttextlabelmatching} are not scalable as they require extra information like label hierarchy. Additionally, some models are built for specific tasks \cite{khatuya-etal-2024-parameter}, making it difficult to generalize to datasets from different domains. 

There also exists popular extreme multi-label classification frameworks like  GalaXC~\cite{saini2021galaxc}, SiameseXML~\cite{dahiya2021siamesexml},  DEXA~\cite{dahiya2023deep}, Renee~\cite{jain2023renee}, InceptionXML~\cite{kharbanda2024inceptionxmllightweightframeworksynchronized} etc. These works focus more on efficiency given the large set of labels to predict from, but rarely utilize label descriptions to enhance performance.

\if{0}
\begin{table}[ht]
 \scriptsize
\begin{tabularx}{\linewidth}{p{2.1cm}|p{1.1cm}|p{2.8cm}}
\hline

\textbf{Text} & \textbf{Labels} &\textbf{Label Descriptions}\\ \hline
\midrule
Maybe I can sleep with my chem book on my head and it will all sink in my brain & pessimism & Pessimism, which includes cynicism and a lack of confidence, is the inclination to expect negative or unfavorable outcomes.\\ \hline
even if it looks like we are okay, the reality of that is we are actually very worn out and forlorn & pessimism, sadness & Pessimism, which includes cynicism and a lack of confidence.... Sadness, encompassing pensiveness and grief, is the feeling of sorrow and melancholy often triggered by loss or unfortunate events.\\ 
\hline
\bottomrule
\end{tabularx}
\caption{Sampled examples of SemEval Dataset showing the importance of label descriptions to classify the text into correct labels. }

\label{tbl:examples}
\end{table}

\noteng{first describe the dataset and then go for data enhancement. 
The data enhancement section is very confusing. IS which label to assign is done using Wikipedia and then assigning that labels to each individual row is done using GPT3.5? }

\fi
\section{Datasets and Label Description Generation}

We curated popular multi-label classification datasets from social media, news, academic, and healthcare domains for our experiments.



\subsection{Overview of Datasets}
\noindent \textbf{CAVES [Social Media]:} The CAVES dataset \cite{poddar2022caves} contains ~10K anti-vaccine tweets related to COVID-19, labeled manually.

\noindent \textbf{Reuters [Newswire]:} Reuters-21578 \cite{Hayes1990CONSTRUETISAS} consists of documents from the 1987 Reuters newswire, with a skewed distribution. 

\noindent \textbf{AAPD [Academic Text]:} The Arxiv Academic Paper Dataset (AAPD) \cite{yang2018sgm} contains abstracts from the computer science domain. 

\noindent \textbf{SemEval [Social Media]:} SemEval-2018 Task 1C \cite{mohammad2018semeval} includes emotion-labeled tweets from 2016-2017 in English. 
 
\noindent \textbf{PubMed [Healthcare]:} A processed version of the PubMed dataset\footnote{\url{https://pubmed.ncbi.nlm.nih.gov/}} from BioASQ 9 Task A\footnote{\url{http://participantsarea.bioasq.org/general_information/Task9a/}}, available on Kaggle\footnote{\url{https://t.ly/FWWuQ}}, manually annotated with Medical Subject Headings (MeSH) by biomedical experts.



\subsection{Dataset Enhancement with Label Descriptions}
Many existing datasets in multi-label classification tasks provide only the final label predictions without offering detailed descriptions for each label. However, label descriptions are essential for improving contextual understanding and enhancing model performance. In this study, we address this gap by augmenting the datasets with refined label descriptions. The CAVES dataset \cite{poddar2022caves}, from the social media domain, is the only dataset in our study that already includes this. 

For the remaining datasets, we generated label descriptions using GPT-3.5 \cite{brown2020language} in combination with predefined Wikipedia definitions. Initially, we retrieved label definitions from Wikipedia, referred to as "initial descriptions." To better align these definitions with the specific context of each dataset, we refined them by providing GPT-3.5 with the initial Wikipedia definition and two relevant examples from the dataset where the label appeared in the predictions. The model was then prompted to generate a more contextually appropriate description for each label, ensuring better alignment with the dataset's context. 

For example, the following prompt was used for the label `\textbf{Anger}' in the SemEval dataset:

\noindent\textit{
\textbf{Label:} Anger \\
\textbf{Initial Description:} Anger, emotion that involves annoyance and rage. \\
\textbf{Dataset:} Contains tweets and corresponding emotion annotations. \\
\textbf{Examples from the dataset:} \\
Tweet 1: "Tears and eyes can dry but I won't, I'm burning like the wire in a lightbulb." \\
Prediction: Anger \\
Tweet 2: "We're going to get City in the next round for a revenge." \\
Prediction: Anger \\
\textbf{Task:} Generate a suitable label description for `Anger` that fits the context of this dataset.
}




\begin{table}[ht]
\centering
\resizebox{\columnwidth}{!}{
\begin{tabular}{l|c|c|c|c|c|c}
\hline
\textbf{Dataset} & \textbf{Train} & \textbf{Dev} & \textbf{Test} & \textbf{\# Labels} & \textbf{Max Labels} & \textbf{Avg. Desc. Length} \\ \hline
CAVES    & 6,957  & 987   & 1,977   & 12  & 3  & 28.17  \\ 
SemEval  & 6,838  & 886   & 3,259   & 12  & 6  & 61.11  \\ 
Reuters  & 6,769  & 1,000 & 3,019   & 90  & 11 & 13.41  \\ 
AAPD     & 53,840 & 1,000 & 1,000   & 54  & 8  & 50.34  \\ 
PubMed   & 40,000 & 10,000& 10,000  & 14  & 13 & 91.40  \\ \hline
\end{tabular}
}
\caption{Dataset statistics. The last three columns show the total number of labels, the maximum number of labels per sample, and the average label description length for each dataset.}
\label{tbl:data_stat}
\end{table}


\noindent \textbf{Evaluation Metrics:} To evaluate the performance of our model, we consider 1) Micro-F1 and 2) Macro-F1 metrics.

\section{Baselines}

We validate our model with different baselines ranging from traditional RNN and CNN based approaches like \noindent{\textbf{TextCNN}} \cite{kim2014convolutional}, \noindent{\textbf{TextRNN}} \cite{liu2016recurrent}, \noindent{\textbf{Attentive ConvNet}}~\cite{yin2018attentive} to transformer based approaches like \noindent{\textbf{BERT}}  \cite{devlin2019bert}, \noindent{\textbf{XLNet}} \cite{yang2020xlnet}, \noindent{\textbf{RoBERTa}} \cite{liu2019roberta}, \noindent{\textbf{StarTransformer}}~\cite{guo2022startransformer}. 
We also compared the performance of \modelname~against various popular extreme multi-label classification frameworks like  \noindent{\textbf{(AttentionXML}}   \cite{you2019attentionxml},  \noindent{\textbf{(GalaXC}}  ~\cite{saini2021galaxc} ,  \noindent{\textbf{SiameseXML}}  \cite{dahiya2021siamesexml},  \noindent{\textbf{DEXA}}   \cite{dahiya2023deep}, \noindent{\textbf{DeepXML}} \cite{dahiya2021deepxml} and  \noindent{\textbf{Renee}}  \cite{jain2023renee}.

We also created our own generative baselines \noindent{\textbf{T5-Base}}~\cite{t5} and \noindent{\textbf{T5-Large}}~\cite{t5}. 
Lastly, with the emergence of \textbf{ChatGPT} \cite{brown2020language}, we were curious to check it’s performance  for this task using the same instruction prompt for 500 random samples using \textit{gpt-3.5-turbo} \footnote{\url{
https://platform.openai.com/docs/models/gpt-3-5}} API.



\begin{figure*}[!thb]
    \centering
    \includegraphics[width=1\linewidth]{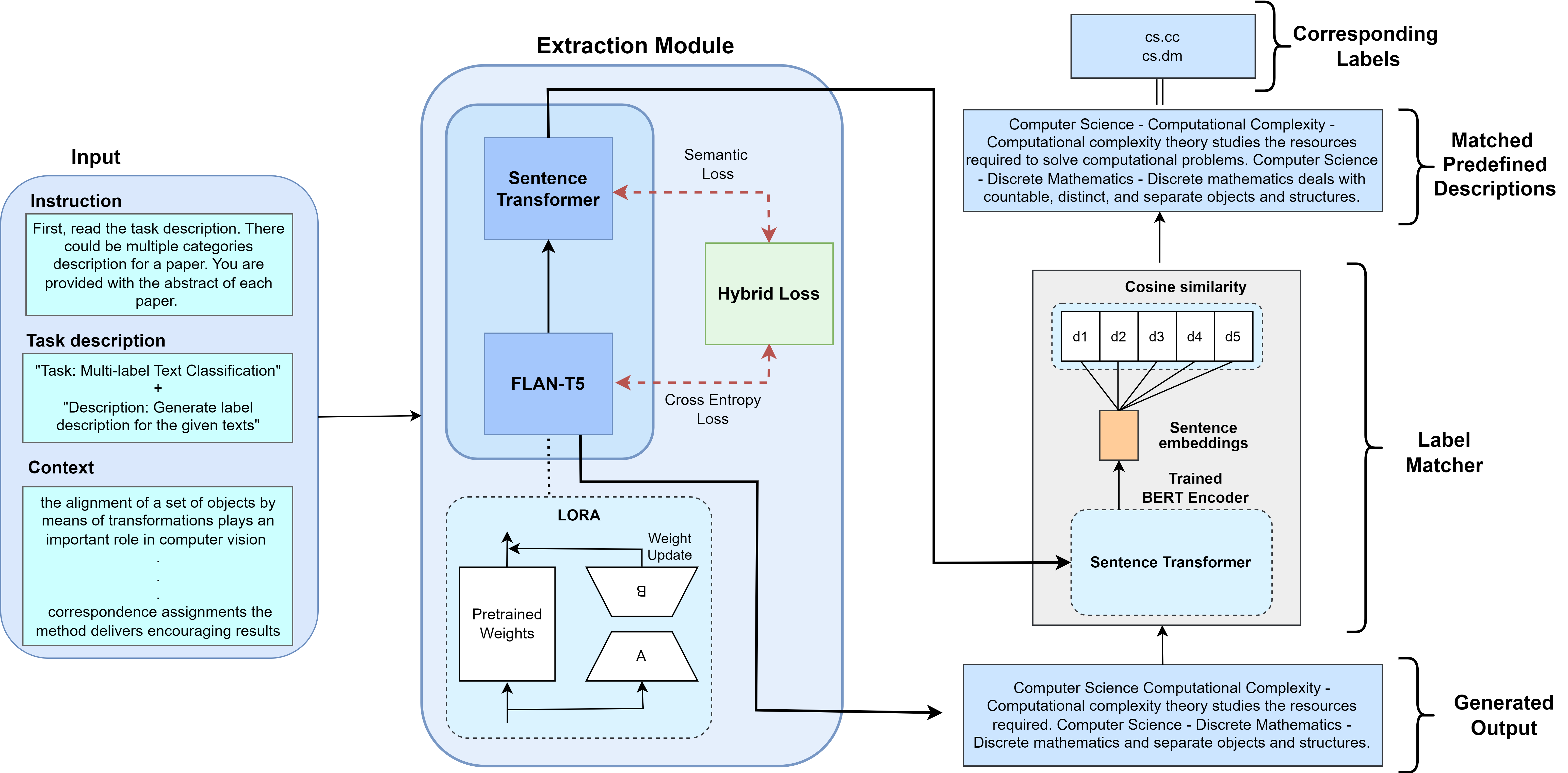}
    \caption{Our proposed framework. Extraction module takes as input a task-specific instruction and the input text to classify. In this module, FLAN-T5 is trained along with a Sentence transformer on a dual objective loss. FLAN-T5-generated label descriptions subsequently flows into the Label Matcher that predicts the final labels for that text using the trained Sentence transformer. }
    \label{fig:arch_diag}


\end{figure*}

\section{Proposed Framework for Multi-label Classification}



The proposed framework \modelname~for multi-label classification (Figure~\ref{fig:arch_diag}), is divided  into two stages:  a supervised generative phase, and an unsupervised description-to-label matching phase.




\noindent\textbf{Problem Formulation:} Given an input sequence $x_{input}$ = \{$x_1$, $x_2$, ..., $x_n$\}, the task is to assign text-class labels $Y_k$ = \{$y_1$, $y_2$, ...., $y_k$\} $\subset Y$ to $x_{input}$ (the text to classify) where  $Y$ = \{$y_1$, $y_2$, ...., $y_p$\} contains all possible labels of that dataset. We adopt prompt-based learning paradigm, generating text conditioned on a given input prompt.

\subsection{Generative Phase}

In the first stage, we frame the problem as a generative task, instructing the model to generate label descriptions from a given document using task-specific prompts. 

\noindent{\textbf{Prompt Construction:}}
We construct a prompt $x_{prompt}$ comprising three components:

\noindent \textbf{Instruction} ($x_{inst}$):  provides a brief overview of input (see Table~\ref{table:instruction_example} for an example).

\noindent \textbf{Task description} ($x_{desc}$):  describes the exact task which needs to be performed. For example in our work when the task the $x_{desc}$ is -\\
\textit{“Task: Multi-label Text Classification \\
Description: Generate label description for the given texts.“}

\noindent \textbf{Input Text} ($x_{input}$): This is the input text sequence which in our case can range from an abstract of a paper to a tweet. 
The $x_{prompt}$ is constructed by concatenation of the Instruction $x_{inst}$, Task Description $x_{desc}$ and Input Text $x_{input}$. An example from the SemEval dataset is given in Table~\ref{table:instruction_example}.

\if{0}
\begin{table}[ht]
\scriptsize

    \begin{tabular}{p{1.5 cm}|p{2.5 cm}}
        \hline
 \textbf{Prompt ($x_{prompt}$)} & \textbf{Target ($y_{target}$)}\\
        \hline 
        \midrule

    
 \parbox[t]{65mm}{
\textbf{Instruction}: First read the task description. There could be multiple categories description for a tweet.\\
\textbf{Task}: Multi-label Text Classification \\
\textbf{Description}: Generate label description for the given texts.

\textit{ It's hot as shit and its fogging up my glasses.} }
&

 \parbox[t]{55mm}{
\textbf{Anger}, which can also encompass annoyance and rage, is a powerful emotion that arises when one feels slighted or wronged. \textbf{Disgust}, which can involve disinterest, dislike, and even loathing, is the strong aversion or revulsion towards something unpleasant or offensive.} \\
\hline
\bottomrule

    \end{tabular}
    
    \caption{Example of prompt (instruction and input text) and target (the label descriptions, separated by full-stop) for the sample  ``\textit{It's hot as shit and its fogging up my glasses}'' having ground truth labels `Anger' and `Disgust' (from the SemEval dataset).}
    \label{table:instruction_example}
\end{table}

\fi

 \begin{table}[ht]
    \centering
    \scriptsize
    \begin{tabular}{p{3.5cm}|p{3.7cm}}
        \hline
        \textbf{Prompt ($x_{prompt}$)} & \textbf{Target ($y_{target}$)} \\
        \hline 
        \midrule
        \parbox[t]{3.5cm}{
        \textbf{Instruction}: First read the task description. There could be multiple categories description for a tweet. \\
        \textbf{Task}: Multi-label Text Classification \\
        \textbf{Description}: Generate label description for the given texts. \\
        \textit{It's hot as shit and its fogging up my glasses.}
        } &
        \parbox[t]{3.8cm}{
        \textbf{Anger}, which can also encompass annoyance and rage, is a powerful emotion that arises when one feels slighted or wronged. \textbf{Disgust}, which can involve disinterest, dislike, and even loathing, is the strong aversion or revulsion towards something unpleasant or offensive.
        } \\
        \hline
        \bottomrule
    \end{tabular}
    
    \caption{Example of prompt (instruction and input text) and target (the label descriptions, separated by a full-stop) from the SemEval dataset.}
    \label{table:instruction_example}
\end{table}

\subsection{Response Construction}
The proposed generative model is expected to generate a textual response $y_{target}$ which is a concatenation of pre-defined label description of the true labels of the corresponding text.
So, if the expected output has $k$ labels = \{$y_1$, $y_2$, ...., $y_k$\} then $y_{target}$ = \{ $y’_1$.$y’_2$........ $y’_k$\} where $y’_i$ denotes the pre-defined label description for the $i^{th}$ label (concatenated and separated using a stop).
Example of $y_{target}$ can be seen in the \textit{Target} column of Table~\ref{table:instruction_example}.
In this generative phase, we formulate $x_{prompt}$, $y_{target}$ for each data point in training dataset. We provide this $x_{prompt}$ as input with target as $y_{target}$ to our model. 

\subsection{Hybrid Loss}
In text generation tasks, models like FLAN-T5 are trained with cross-entropy (CE) loss. Cross-entropy loss operates at a token level, meaning it only rewards exact matches at each position in the sequence. The primary limitation arises from its inability to account for semantically equivalent sentences that use different tokens. 
Hence, we incorporate a semantic similarity based term in the loss function while training the generative model, which prevents the model from overfitting to exact matches. Using such a hybrid loss helps bring the embeddings of the outputs of the generative model and the embeddings of the predefined label description closer in the representation space, making it easier to map the generated outputs with final labels. We define the hybrid loss function as follows:

\begin{equation}
\mathcal{L}_{\text{hybrid}} = \lambda \cdot \mathcal{L}_{\text{CE}} + (1 - \lambda) \cdot \mathcal{L}_{\text{semantic}}
\end{equation}

where: $\mathcal{L}_{\text{CE}} = - \sum_{t=1}^{T} y_t \log(\hat{y}_t)$ represents the traditional cross-entropy loss, which is computed at the token level. Here, $y_t$ is the ground-truth token at position $t$, and $\hat{y}_t$ is the predicted probability for the token at the same position.
    
 $\mathcal{L}_{\text{semantic}} = 1 - \text{CosSim}(v_{\text{gen}}, v_{\text{target}})$ is the semantic similarity loss, where $v_{\text{gen}}$ and $v_{\text{target}}$ represent the embeddings produced by the sentence transformer for the generated output from generative model and the target $y_{target}$, respectively. $CosSim$ denotes the Cosine similarity between the generated and target embeddings.
    The \textit{sentence transformer is allowed to train} and adapt during the learning process. $\lambda$ is a learnable parameter that dynamically adjusts the balance between cross-entropy loss and semantic similarity loss.
    

\subsection{Label Matching Phase}
\label{label matcher section}

During inference, we employ a \textit{Label Matcher} module to assign labels based on similarity between generated and predefined descriptions. We utilize the \textit{trained sentence transformer from the generative phase} to obtain embeddings for both the generated sentences $\{gendesc_1, gendesc_2, \dots, gendesc_k\}$ and the predefined label descriptions. For each generated sentence $gendesc_i$, we compute its cosine similarity with all label embeddings and select the label with the highest similarity as the final prediction $predLabel_i$. This approach ensures robust matching, even when the generated descriptions deviate from the predefined labels.

\subsection{Generative Models Explored} 
To evaluate the viability of a generative approach for this task, we conduct a comprehensive assessment of multiple generative models, varying in size and training strategies. First, we fine-tune T5-Base (220M parameters) and T5-Large (780M parameters), to generate target descriptions. Next, we fine-tune FLAN-T5 Large~\cite{longpre2023flan,chung2022scaling}, which benefits from extensive pre-training on over 1.8K instruction-based tasks. For efficient fine-tuning, we apply Low-Rank Adaptation (LoRA)~\cite{hu2021lora} for all the generative models updating just 0.08\% of model parameters. 
Details on trainable parameters for our models and baselines are provided in Table \ref{table:results_table}.

\section{Experimental Setup}



For all our model variants (performed on NVIDIA A100 80G GPUs),
we obtain the pre-trained checkpoints from the \textit{Huggingface} Library\footnote{\url{
https://huggingface.co/}}.
 For training the models with LoRA, the \textit{rank} for the trainable decomposition matrices was set to 2. FLAN-T5-Large model is instruction-tuned for 20 epochs, with batch-size of 8 and with an lr of $2e-4$ with LoRA (training time: 56 minutes/epoch, inference time: 2 minutes/sample).
These hyperparameters were selected based on the best Macro-F1 results on the validation set. The input length was set by the average number of input tokens per dataset, while the output length was based on the average label description length for each dataset.


\if{0}
\begin{table*}[tb]
\scriptsize
\centering
\begin{tabular}{l|cc|cc|cc|cc|cc} 
\hline
\multicolumn{1}{c}{} & \multicolumn{2}{c}{\textbf{CAVES}} & \multicolumn{2}{c}{\textbf{SemEval}} & \multicolumn{2}{c}{\textbf{Reuters}} & \multicolumn{2}{c}{\textbf{AAPD}} & \multicolumn{2}{c}{\textbf{PubMed}} \\ 
\hline
\textbf{Models}  & \textbf{Mi-F1} & \textbf{M-F1} & \textbf{Mi-F1} & \textbf{M-F1}  & \textbf{Mi-F1} & \textbf{M-F1} & \textbf{Mi-F1} & \textbf{M-F1}  & \textbf{Mi-F1} & \textbf{M-F1} \\ 
\hline
\textbf{Baselines} &  &  &  &  &  &  &  &  &  &  \\ 
\hline
BERT  & 70.36 & \underline{65.29}  & \underline{70.70} & \underline{56.30}  & 87.73 & 34.98  & \underline{71.30} & 55.90  & 85.05 & \underline{70.99} \\
XLNet  & \underline{71.61} & 63.83 & 58.01 & 35.31 & \textbf{88.54} & \underline{51.99} & 70.07 & \underline{58.39}  & \underline{85.33} & 70.81 \\
RoBERTa  & 71.34 & 63.82  & 59.82 & 40.55  & 88.27 & 42.63 & 69.14 & 54.88  & 85.19 & 70.54 \\
TextCNN  & 55.48 & 39.64 & 54.55 & 39.51 & 81.89 & 33.96  & 67.71 & 49.85 & 82.62 & 66.4 \\
TextRNN  & 57.70 & 42.17  & 52.42 & 37.94 & 81.72 & 33.26  & 69.28 & 52.27  & 83.11 & 67.72 \\
StarTransformer  & 53.86 & 35.98 & 51.42 & 38.96  & 80.22 & 36.39  & 68.22 & 49.36 & 82.35 & 67.35 \\
AttentiveConvNet & 54.22 & 38.15  & 51.61 & 37.21  & 79.77 & 31.86 & 68.11 & 49.21  & 82.65 & 66.33 \\
ChatGPT (500 samples) & 57.22 & 44.47  & 41.61 & 27.21  & 69.77 & 35.86 & 48.11 & 29.21  & 42.65 & 26.33 \\
AttentionXML & - & -  & - & -  & - & - & - & -  & - & - \\

GalaXC & - & -  & - & -  & - & - & - & -  & - & - \\

SiameseXML & - & -  & - & -  & - & - & - & -  & - & - \\

DEXA & - & -  & - & -  & - & - & - & -  & - & - \\

DeepXML & - & -  & - & -  & - & - & - & -  & - & - \\

Renee & - & -  & - & -  & - & - & - & -  & - & - \\

\hline
\textbf{Proposed Models} &  &  &  &  &  &  &  &  &  &   \\ 
\hline
T5-Base  & 86.84 & 71.22 & 83.12 & 70.13 & 87.89 & 55.23 & 82.25 & 71.22 & 85.87 & 72.26  \\
T5-Large  & 88.33 & 81.89 & 84.35 & 72.23 & 88.12 & 57.32 & 84.23 & 71.59 & 86.25 & 73.35  \\
\textbf{Ours}  & \textbf{89.67} & \textbf{85.25}  & \textbf{85.53} & \textbf{74.79}  & 88.20 & \textbf{58.56}  & \textbf{86.94} & \textbf{73.13} & \textbf{86.77} & \textbf{74.12} \\

\textbf{Ours with emb Loss}  & \textbf{92.46} & \textbf{89.11}  & \textbf{87.81} & \textbf{78.06}  & \textbf{96.48} & \textbf{80.85}  & \textbf{-} & \textbf{-} & \textbf{89.93} & \textbf{80.81} \\
\hline
\end{tabular}
\caption{Performance evaluation based on Micro F1 and Macro F1. The best performance is highlighted in bold, and the strongest baseline result is underlined.}
\label{table:results_table}
\end{table*}

\fi

\begin{table*}[ht]
\centering
 \scalebox{0.85}
 {
\begin{tabular}{p{2.52cm}|p{1.1cm}p{0.95cm}|cc|cc|cc|cc|p{0.10cm}} 
\hline
\multicolumn{1}{c}{} & \multicolumn{2}{c}{\textbf{CAVES}} & \multicolumn{2}{c}{\textbf{SemEval}} & \multicolumn{2}{c}{\textbf{Reuters}} & \multicolumn{2}{c}{\textbf{AAPD}} & \multicolumn{2}{c|}{\textbf{PubMed}} & \textbf{\#TP(M)} \\ 
\hline
\textbf{Models}  & \textbf{Mi-F1} & \textbf{M-F1} & \textbf{Mi-F1} & \textbf{M-F1}  & \textbf{Mi-F1} & \textbf{M-F1} & \textbf{Mi-F1} & \textbf{M-F1}  & \textbf{Mi-F1} & \textbf{M-F1} &  \\ 
\hline
\textbf{Baselines} &  &  &  &  &  &  &  &  &  &  &  \\ 
\hline
BERT  & 70.36 & 65.29  & 70.70 & 56.30  & 87.73 & 34.98  & 71.30 & 55.90  & 85.05 & 70.99 & 110 \\ 
XLNet  & 71.61 & 63.83 & 58.01 & 35.31 & 88.54 & 51.99 & 70.07 & 58.39  & 85.33 & 70.81 & 110 \\ 
RoBERTa  & 71.34 & 63.82  & 59.82 & 40.55  & 88.27 & 42.63 & 69.14 & 54.88  & 85.19 & 70.54 & 125 \\ 
TextCNN  & 55.48 & 39.64 & 54.55 & 39.51 & 81.89 & 33.96  & 67.71 & 49.85 & 82.62 & 66.4 & 3.88 \\ 
TextRNN  & 57.70 & 42.17  & 52.42 & 37.94 & 81.72 & 33.26  & 69.28 & 52.27  & 83.11 & 67.72 & 3.86 \\ 
StarTransformer  & 53.86 & 35.98 & 51.42 & 38.96  & 80.22 & 36.39  & 68.22 & 49.36 & 82.35 & 67.35 & 3.84 \\ 
AttentiveConvNet & 54.22 & 38.15  & 51.61 & 37.21  & 79.77 & 31.86 & 68.11 & 49.21  & 82.65 & 66.33 & 3.91 \\ 
ChatGPT & 57.22 & 44.47  & 41.61 & 27.21  & 69.77 & 35.86 & 48.11 & 29.21  & 42.65 & 26.33 & - \\ 
AttentionXML & 67.12 & 52.83  & 61.55 & 48.32  & 78.43 & 43.57 & 69.01 & 56.28  & 84.39 & 69.11 & 112 \\ 

GalaXC & 69.84 & 56.12  & 62.34 & 50.79  & 81.23 & 46.39 & 70.65 & 58.17  & 85.91 & 71.34 & 41 \\ 

SiameseXML & 72.16 & 59.89  & 65.42 & 52.84  & 84.22 & 48.99 & 72.48 & 60.43  & 86.42 & 72.98 & 115 \\ 

DEXA & 74.81 & 62.35  & 66.57 & 54.32  & 86.07 & 51.12 & 74.21 & 63.12  & 87.25 & 74.22 & 134 \\ 

DeepXML & 77.53 & 65.84  & 68.76 & 55.98  & 88.52 & 53.79 & 75.63 & 64.58  & 88.41 & 75.96 & 161 \\ 

Renee & 79.46 & 67.93  & 71.18 & 58.43  & 90.29 & 66.21 & 77.82 & 66.04  & 89.74 & 78.12 & 82 \\ 

\hline
\textbf{Proposed} &  &  &  &  &  &  &  &  &  &  &  \\ 
\hline
T5-Base  & 86.84 & 71.22 & 83.12 & 70.13 & 90.89 & 69.23 & 82.25 & 71.22 & 89.87 & 78.56 & 22.32 \\ 
T5-Large  & 88.33 & 81.89 & 84.35 & 72.23 & 91.12 & 71.32 & 84.23 & 71.59 & 89.90 & 79.35 & 22.68 \\

\textbf{\modelname}  & \textbf{92.46} & \textbf{89.11}  & \textbf{87.81} & \textbf{78.06}  & \textbf{96.48} & \textbf{80.85}  & \textbf{95.64} & \textbf{88.43} & \textbf{89.93} & \textbf{80.81} & 22.69 \\ 
\hline
\end{tabular}
 }


\caption{Performance evaluation based on Micro F1 and Macro F1 scores across multiple datasets. The best performance is highlighted in bold, and the strongest baseline result is underlined. The last column TP (M) indicate approx. no of trainable parameters in million.}
\label{table:results_table}
\end{table*}

\section{Results and Discussion}
 We report the results of our proposed generative model variants and various baselines in Table \ref{table:results_table} for all the datasets. We first finetune pretrained transformers such as BERT, XLNet, RoBERTa with a projection layer for the task of MLTC. Next we compare with baselines designed specifically for the task of MLTC such as GalaXC, DeepXML, Renee, etc. We observe that among the baselines the models designed specifically for MLTC task outperform the finetuned transformers. To check the robustness of our proposed generative framework, we try out with different base models such as T5-Base, T5-Large. Our devised generative baselines outperform the best baselines across all datasets. We also report ChatGPT's performance by providing the same prompt as provided to \modelname.

\modelname~on an average improves by \textbf{~13.94\%} in Micro-F1 and \textbf{~24.85\% }in Macro-F1 when compared with the best baseline for the given dataset. The best performance boost is seen for SemEval \footnote{Public leaderboard available at \url{https://paperswithcode.com/sota/emotion-classification-on-semeval-2018-task}} and the second best for the CAVES dataset. The performance of \modelname~compared to the respective SOTA models in domains ranging from tweet sentiment to medical domain dataset, academic text showcases its adaptability and versatility. 

\subsection{Parameter Efficiency}
The last column of Table~\ref{table:results_table} compares the number of trainable parameters across different baselines, including the generative models we trained. Our best-performing model, \modelname~ along with our generative baselines, demonstrates parameter efficiency by having significantly fewer trainable parameters compared to most of the closest competing baselines.


\subsection{Utility of Label Descriptions}
To understand the importance of label descriptions, we perform an experiment where we set atomic labels instead of their descriptions, as the target. Accordingly, the \textit{Label Matcher} module now compares the embeddings of the generated and ground truth labels (and not descriptions).
From the Table~\ref{tbl:imp_label_desc_semantic_loss}, we observe a average drop of \textbf{36.04\%} in Macro-F1. 

\subsection{Utility of Semantic Loss}
We assess the significance of semantic loss by comparing the performance of our proposed hybrid loss function with that of the standard cross-entropy loss. The results, summarized in Table~\ref{tbl:imp_label_desc_semantic_loss}, show a drop of $\textbf{3.96}\textbf{\%}$ and $\textbf{7.33}\textbf{\%}$ in Micro-F1 and Macro-F1, respectively, when using only cross-entropy loss. 
\section{Analysis}
We now present different analyses and ablations of \modelname.
\subsection{Zero Shot Capability}

\begin{figure}[ht]

  \centering
  \includegraphics[width=\linewidth]{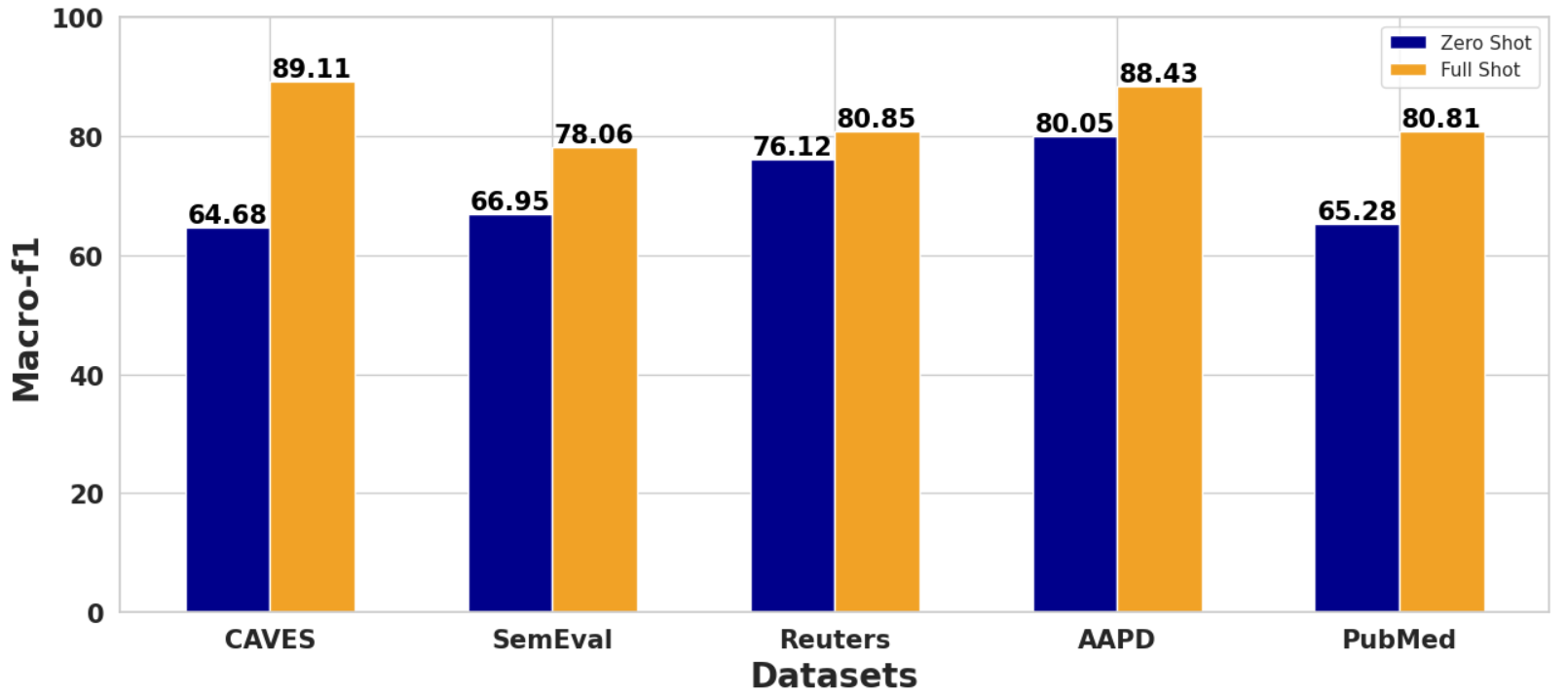}


\caption{Zero-shot performance of \modelname~, achieving an average macro-F1 score of 70.61.}



\label{fig:zero_shot_performance}
\end{figure}

To evaluate the zero shot capability of our proposed model, we constructed a test dataset with labels not seen during training. Specifically, 4-5 labels from each dataset were randomly selected as unseen labels, appearing only in test instances. We then trained the model on the modified dataset to assess its ability to predict unseen labels. As shown in Figure~\ref{fig:zero_shot_performance}, the average Macro-F1 score across all datasets was 83.45 with full training, compared to 70.61 in the zero-shot setting—demonstrating strong performance in the more challenging zero-shot scenario. This reduces the need for retraining and also accelerates real-world deployment.


\subsection{Performance on Least Frequent Labels}

\begin{figure}[ht]

  \centering
  \includegraphics[width=0.92\linewidth]{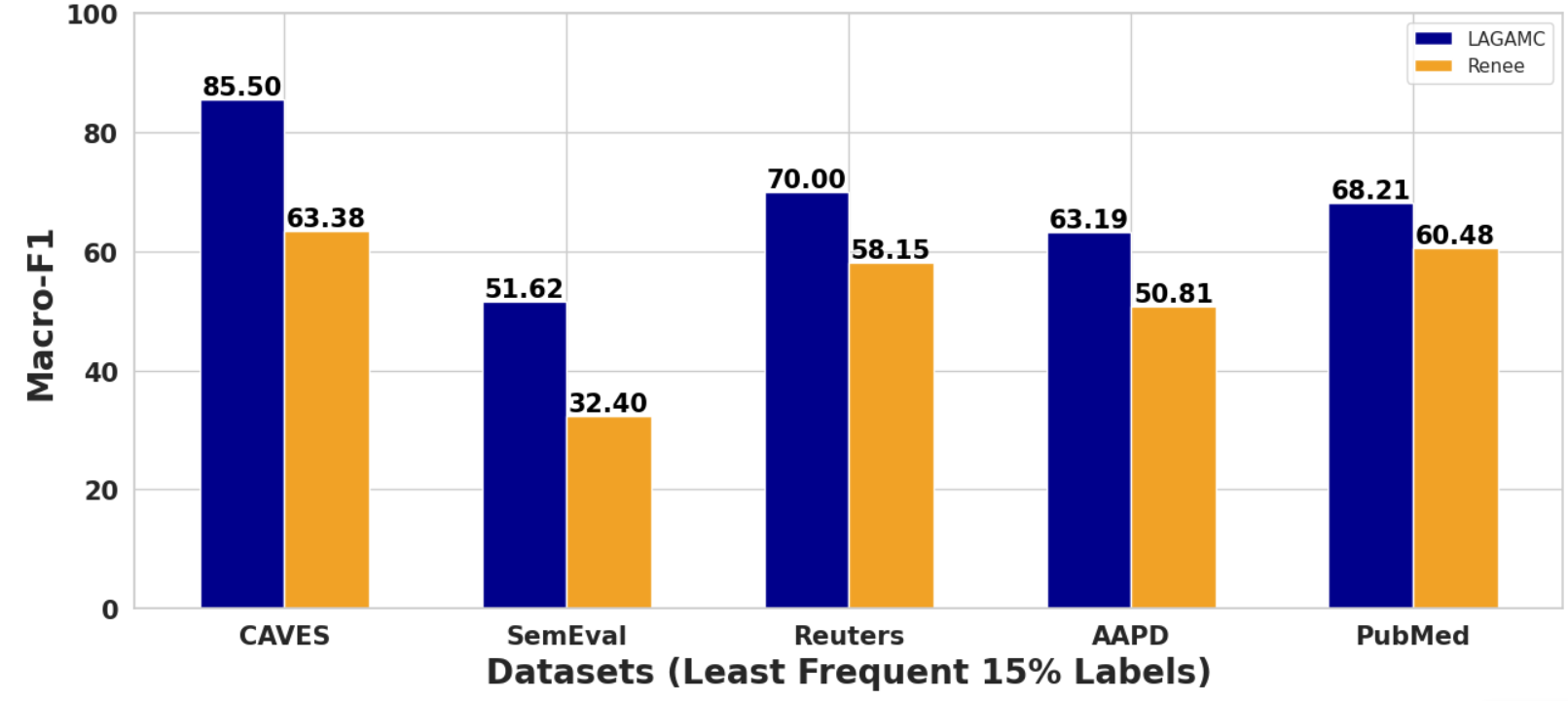}

\caption{Comparison of Model Performance on Least Frequent Labels: Our proposed model demonstrates superior performance compared to closest baseline. }

\label{fig:least_freq_performance}
\end{figure}

We evaluate the model's performance on the least frequent labels as this is critical for real-world applications, where rare labels may represent significant events. We identified the least frequent 15\% of labels from each dataset's training set and computed the Macro F1-score on test samples where the ground truth labels were part of this rare label set. As illustrated in Figure~\ref{fig:least_freq_performance}, our model demonstrates superior performance over the closest baseline across all datasets, with an average improvement of \textbf{22\%} in Macro-f1.

\subsection{Evaluation of Recent LLM's}
We also evaluated recent LLMs, such as Llama-2-7b \cite{llama2}, Mistral-7B \cite{mistral7b} for multi-label classification. The initial results shown in Table \ref{tab:llm_results} were promising, as these models outperformed all baseline methods (except Micro-F1 on PubMed which is close to the best baselines). However, their accuracy was lower than our proposed \modelname, which uses FLAN-T5. Due to limited GPU resources, we could not fully fine-tune these models or conduct an extensive hyperparameter search for LoRA fine-tuning, which likely contributed to the lower performance. We also expect that increasing the context length could improve results. These findings indicate that our pipeline is effective and can be applied to other LLMs for multi-label classification. We also evaluated our pipeline with Llama-3.1-7B \cite{grattafiori2024llama3herdmodels} and observed an improvement of nearly 1\% in Micro-F1 and 2\% in Macro-F1.

\begin{table}[ht]
\centering
\scriptsize
\resizebox{\linewidth}{!}{
\begin{tabular}{l|l c}
\hline
\textbf{Dataset} & \textbf{Model} & \textbf{Mi-F1 / M-F1} \\ 
\hline
\multirow{4}{*}{CAVES}
& Ours & 92.46 / 89.11 \\
& Ours with threshold & 91.46 / 88.55 \\
& Llama-2-7B with threshold & 90.97 / 87.13 \\
& Llama-2-7B w/o threshold & 89.67 / 83.65 \\
& Mistral-7B with threshold & 90.17 / 86.52 \\
& Mistral-7B w/o threshold & 88.59 / 82.25 \\
\hline
\multirow{4}{*}{SemEval} 
& Ours & 87.81 / 78.06 \\
& Ours with threshold & 86.18 / 77.60 \\
& Llama-2-7B with threshold & 86.54 / 77.11 \\
& Llama-2-7B w/o threshold & 84.67 / 75.65 \\
& Mistral-7B with threshold & 85.54 / 76.25 \\
& Mistral-7B w/o threshold & 84.45 / 74.45 \\
\hline
\multirow{4}{*}{Reuters} 
& Ours & 96.48 / 80.85 \\
& Ours with threshold & 94.48 / 76.15 \\
& Llama-2-7B with threshold & 93.62 / 74.65 \\
& Llama-2-7B w/o threshold & 92.18 / 72.88 \\
& Mistral-7B with threshold & 95.17 / 76.43 \\
& Mistral-7B w/o threshold & 93.55 / 74.15 \\
\hline
\multirow{4}{*}{AAPD} 
& Ours & 95.64 / 88.43 \\
& Ours with threshold & 94.46 / 86.73 \\
& Llama-2-7B with threshold & 93.46 / 86.11 \\
& Llama-2-7B w/o threshold & 89.12 / 75.97 \\
& Mistral-7B with threshold & 92.76 / 86.07 \\
& Mistral-7B w/o threshold & 87.57 / 74.15 \\
\hline
\multirow{4}{*}{PubMed} 
& Ours & 89.93 / 80.81 \\
& Ours with threshold & 89.22 / 78.91 \\
& Llama-2-7B with threshold & 89.92 / 79.71 \\
& Llama-2-7B w/o threshold & 86.75 / 77.45 \\
& Mistral-7B with threshold & 89.67 / 79.55 \\
& Mistral-7B w/o threshold & 87.65 / 78.35 \\
\hline
\end{tabular}
}
\caption{Performance comparison (Mi-F1 / M-F1) on multiple datasets using different LLMs.}
\label{tab:llm_results}
\end{table}

\subsection{Robustness of the Model}
To assess the robustness of our approach, we created a unified dataset by randomly selecting 500 training samples and 100 test samples from each dataset. The average Macro-F1 score was 83.45 across all datasets, compared to 78.38 for the mixed dataset. Despite predicting from 181 ground truth labels (sum of Labels column in Table~\ref{tbl:data_stat}), our model shows only a 5.07\% drop in performance, while the closest baseline declines by 14.3\% .

\begin{table}[ht]
\resizebox{0.9\linewidth}{!}{
\begin{tabular}{l|cc}
\hline
\textbf{Dataset} & \textbf{Models} & \textbf{Mi-F1 / M-F1} \\ 
\hline
\multirow{3}{*}{CAVES} 
& Ours & 92.46 / 89.11 \\
& w/o Semantic loss & 89.67 / 85.25\\
& w/o Label Description & 67.63 / 62.98 \\
\hline
\multirow{3}{*}{SemEval} 
& Ours & 87.81 / 78.06 \\
& w/o Semantic loss & 85.53 / 74.79  \\
& w/o Label Description & 64.24 / 54.35 \\
\hline
\multirow{3}{*}{Reuters} 
& Ours & 96.48 / 80.85 \\
& w/o Semantic loss & 94.97 / 78.12 \\
& w/o Label Description & 68.24 / 43.12 \\
\hline
\multirow{3}{*}{AAPD} 
& Ours & 95.64 / 88.43 \\
& w/o Semantic loss & 86.94 / 73.13 \\
& w/o Label Description & 65.27 / 53.28 \\
\hline
\multirow{3}{*}{PubMed} 
& Ours & 89.93 / 80.81 \\
& w/o Semantic loss & 86.77 / 74.12 \\
& w/o Label Description & 67.29 / 53.21 \\
\hline
\end{tabular}
}
\caption{Performance comparison with, without label descriptions and without Semantic loss across datasets. Results highlight the importance of label descriptions and Semantic loss for multi-label classification.}
\label{tbl:imp_label_desc_semantic_loss}
\end{table}

\begin{figure}[ht]
\centering
\begin{minipage}{0.50\linewidth}
  \centering
  \includegraphics[width=.95\linewidth]{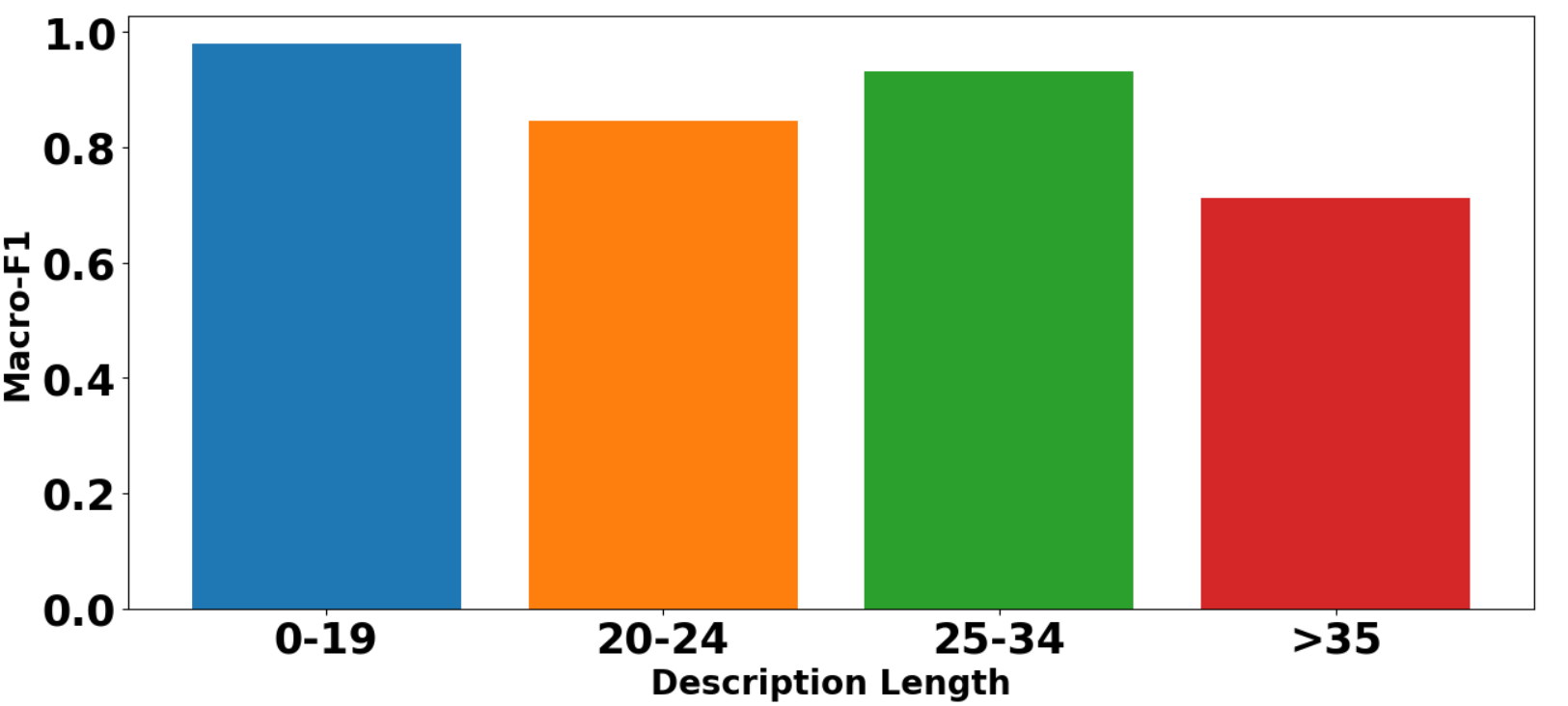}
  \caption{CAVES}
  \label{fig:CAVES}
\end{minipage}%
\hfill
\begin{minipage}{0.50\linewidth}
   \centering
  \includegraphics[width=.95\linewidth]{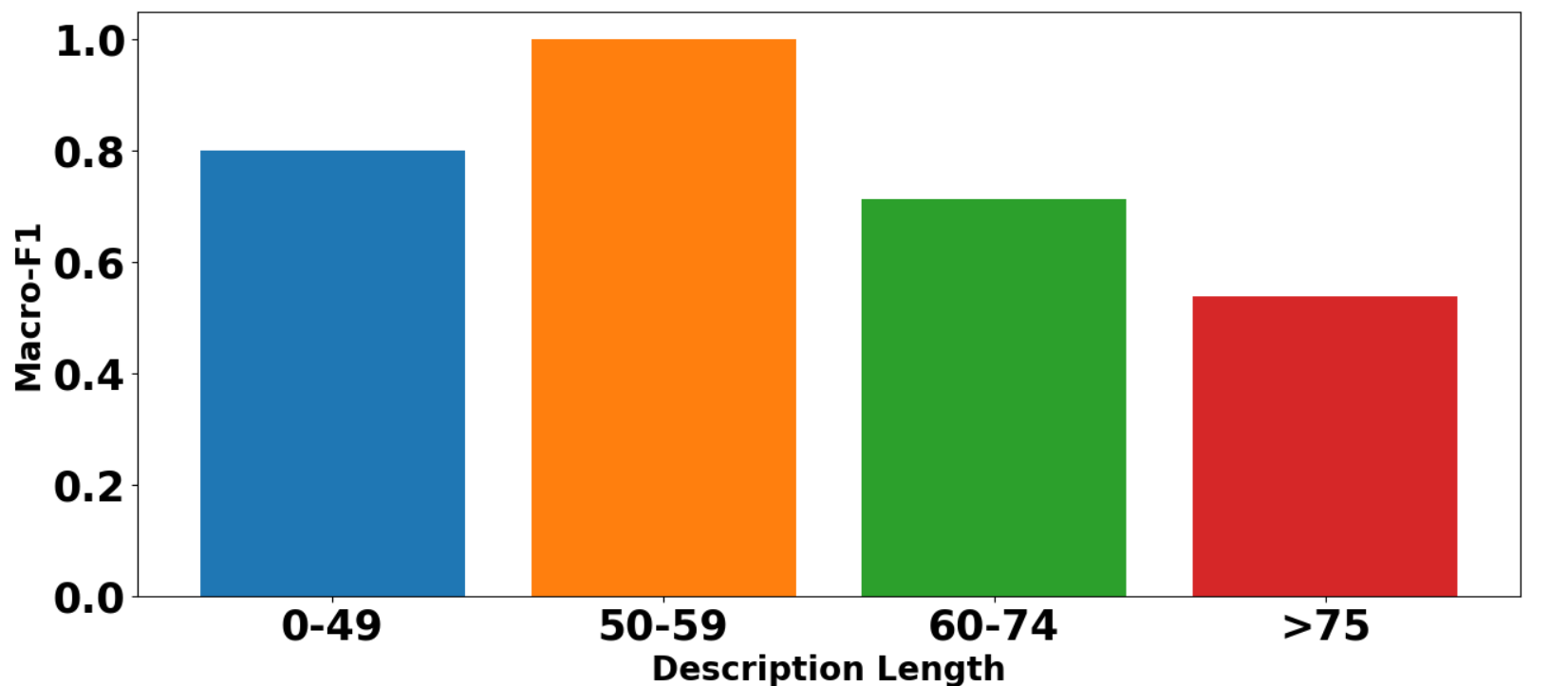}
  \caption{SemEval}
  \label{fig:Semeval}
\end{minipage}
\caption{Impact of label description length}
\label{fig:length_performance}
\end{figure}

\noindent \textbf{Descriptions Length vs Performance:}
 We evaluate our model's performance based on the length of concatenated label descriptions. To do this, we group label descriptions into buckets with an equal number of test samples. Longer descriptions correspond to more ground truth labels, increasing prediction complexity. Figures~\ref{fig:CAVES} and \ref{fig:Semeval} show a slight performance drop for very long descriptions, a trend consistent across datasets.

\noindent \textbf{Actual vs Predicted No of Labels:}
For each dataset, we analyze the number of samples with a given number of labels and compare it to the number of samples predicted to have the same number of labels (Table ~\ref{tbl:actual_vs_predicted_labels}). We have analyzed upto five label counts. The model tends to give single-label predictions for the SemEval and Caves datasets.

\begin{table}[ht]
\centering

\resizebox{\linewidth}{!}{
\begin{tabular}{c|c|c|c|c|c}
\hline
\textbf{Labels} & \textbf{Caves} & \textbf{SemEval} & \textbf{Reuters} & \textbf{AAPD} & \textbf{PUBMED} \\ 
\hline
1 & (1386, 1562) & (288, 456) & (2592, 2583) & (-, -) & (-, -) \\ 
2 & (579, 369) & (1486, 1367) & (279, 308) & (642, 690) & (50, 105) \\ 
3 & (12, 46)   & (1078, 1055) & (86, 63)   & (264, 225) & (376, 535) \\ 
4 & (-, -)     & (395, 316)  & (32, 32)   & (69, 62)   & (1438, 1503) \\ 
5 & (-, -)     & (11, 60)    & (17, 15)   & (23, 21)   & (2200, 2404) \\ 

\hline
\end{tabular}
}
\caption{Comparison of actual and predicted sample counts based on number of labels. Each cell (x, y) indicates the number of actual samples (x) and the number of predicted samples (y) for a specific label count.}
\label{tbl:actual_vs_predicted_labels}

\end{table}


\subsection{Analysis of Label Matcher Module}  
We examine the computational efficiency of the module and propose a threshold-based approach to prevent label assignments from hallucinated text.


\noindent \textbf{Computation Efficiency:} The Label Matcher module (Section~\ref{label matcher section}) assigns labels by computing cosine similarity between sentence and label embeddings. Using NumPy’s~\cite{numpy} matrix operations for parallel computation significantly improves efficiency. For example, with 10,000 sentences and 1,000 labels using 1,024-dimensional embeddings, the matrix-based approach completes in 0.089s, compared to 0.354s with the sequential method. In the worst-case with all 1,000 labels present in a single instance, inference takes just 0.007s (matrix-based) versus 0.043s (sequential).

\noindent \textbf{Hallucination in Predictions:} During label matching, each output sentence is assigned the nearest label based on cosine similarity. However, LLM-generated text may include hallucinated sentences, leading to incorrect predictions. To mitigate this, we enforce a minimum similarity threshold, ensuring a sentence is assigned a label only if its highest similarity score exceeds a set value. Our analysis finds 0.4 threshold to be optimal. As shown in Table \ref{tab:llm_results}, this slightly reduces performance for our model (FLAN-T5-based) but improves results for larger LLMs like Llama-2 and Mistral, likely due to their tendency to generate longer outputs.

\if{0}
\begin{table}[ht]
\centering
\scalebox{0.7}{
\begin{tabular}{l|cc}
\hline
\textbf{Dataset} & \textbf{Models} & \textbf{Mi-F1 / M-F1} \\ 
\hline
\multirow{4}{*}{CAVES} 
& Ours & 92.46 / 89.11 \\
& w/ S-BERT-L12 & 87.20 / 82.80 \\
& w/ ST5-xxl & 87.00 / 83.20 \\
& w/o instruction & 86.00 / 81.30 \\
\hline
\multirow{4}{*}{SemEval} 
& Ours & 87.81 / 78.06 \\
& w/ S-BERT-L12 & 85.00 / 74.30 \\
& w/ ST5-xxl & 85.30 / 74.70 \\
& w/o instruction & 85.20 / 74.80 \\
\hline
\multirow{4}{*}{Reuters} 
& Ours & 96.48 / 80.85 \\
& w/ S-BERT-L12 & 88.20 / 58.60 \\
& w/ ST5-xxl & 88.50 / 56.70 \\
& w/o instruction & 88.20 / 56.40 \\
\hline
\multirow{4}{*}{AAPD} 
& Ours & 95.64 / 88.43 \\
& w/ S-BERT-L12 & 86.90 / 71.40 \\
& w/ ST5-xxl & 86.50 / 71.40 \\
& w/o instruction & 86.70 / 71.60 \\
\hline
\multirow{4}{*}{PubMed} 
& Ours & 89.93 / 80.81 \\
& w/ S-BERT-L12 & 86.80 / 74.10 \\
& w/ ST5-xxl & 87.70 / 75.00 \\
& w/o instruction & 87.00 / 74.30 \\
\hline
\end{tabular}
}
\caption{Ablation results: Mi-F1 and M-F1 scores for \modelname~and variations with label matcher modules, without instruction.}



\label{tbl:ablation_study}
\end{table}

\fi


\begin{table}[ht]
\centering
\renewcommand{\arraystretch}{1.2}
\scalebox{0.65}{
\begin{tabular}{l|c|c|c|c|c}
\hline
\textbf{Model} & \textbf{CAVES} & \textbf{SemEval} & \textbf{Reuters} & \textbf{AAPD} & \textbf{PubMed} \\
\textbf{} & (M-F1) & (M-F1) & (M-F1) & (M-F1) & (M-F1) \\\hline
Ours                & 89.11 & 78.06 & 80.85 & 88.43 & 80.81 \\
\quad w S-BERT-L12         & 82.80 & 74.30 & 58.60 & 71.40 & 74.10 \\
\quad w ST5-xxl            & 83.20 & 74.70 & 56.70 & 71.40 & 75.00 \\
w/o Instruction     & 81.30 & 74.80 & 56.40 & 71.60 & 74.30 \\\hline
\end{tabular}
}
\caption{Ablation study results: M-F1 scores for \modelname~and its variations. }
\label{tbl:ablation_study}
\end{table}

\subsection{Ablation study of model components} 

We conduct ablations on our best model to assess module significance. For \textit{Label Matcher}, replacing fine-tuned Sentence-BERT-Transformer with Sentence-T5-xxl or Sentence-BERT-L12~\cite{reimers2019sentence} lowers performance (Table~\ref{tbl:ablation_study}). Similarly, instruction-tuning FLAN-T5-Large \textit{without task-specific instructions} results in a performance drop across all datasets, highlighting the importance of instruction alignment.


\subsection{Effect of Label Descriptions in Existing Models}
 We conducted additional experiments by integrating label descriptions into two strong baselines: BERT and DeepXML.
For the BERT-based multi-label classification baseline, we adopted a joint encoding strategy where both the input text and the label descriptions are encoded using a shared BERT encoder. This allows the model to learn interactions between label semantics and text representations. 

For DeepXML, which supports metadata incorporation, we introduced label descriptions as auxiliary features. Additionally, during the clustering phase, we replaced label names with their corresponding descriptions to influence label partitioning based on semantic content.

We evaluated these modified models on the CAVES and SemEval datasets. The results are presented in Table~\ref{tab:label-desc-comparison}. We observe that incorporating label descriptions provides consistent but modest performance improvements over the original versions of BERT and DeepXML. However, our proposed generative approach with the label matcher module achieves significantly better performance, demonstrating the advantage of a design that integrates label semantics into the prediction process.

\begin{table}[b]
\centering
\scriptsize
\setlength{\tabcolsep}{3pt}

\begin{tabular}{lcc}
\toprule
\textbf{Model} & \textbf{CAVES Mi-F1 / M-F1} & \textbf{SemEval Mi-F1 / M-F1} \\
\midrule
BERT & 70.36 / 65.29 & 70.70 / 56.30 \\
BERT + Label Desc. & 73.00 / 67.50 & 72.20 / 58.00 \\
DeepXML & 77.53 / 65.84 & 68.76 / 55.98 \\
DeepXML + Label Desc. & 79.50 / 67.20 & 70.12 / 57.30 \\
\bottomrule
\end{tabular}
\caption{Performance of baselines with and without label descriptions.}
\label{tab:label-desc-comparison}
\end{table}

\subsection{Error Analysis}
\label{subsec:error_analysis}

To characterize the errors committed by our model, we check when our model predicts wrong label or provides a subset of ground truth labels.  
We notice that our model sometime struggles against complex inputs. The first example stated in Table \ref{tbl:error-analysis} is about a news related to a company dealing with gold but the news excerpt is regarding acquisition of it. This confuses our model because even though the word gold is mentioned multiple times, the main subject of the news is regarding the acquisition rather than about gold commodities.

\begin{table}[ht]
\centering
\scriptsize
\begin{tabularx}{\linewidth}{p{0.8cm}|p{2.0cm}|p{1 cm}|p{1 cm}|p{1.1cm}}
\hline
\textbf{Dataset} & \textbf{Abstract} & \textbf{Ground Truth} & \textbf{Ours} & \textbf{Renee} \\ 
\hline
Reuters & CRA SOLD FORREST GOLD FOR 76 MLN DLRS... It also owns an undeveloped gold project. & acq & acq, gold & acq, gold \\ 
\hline
CAVES & The covid vaccine is not a vaccine... the next round of manufactured flu. & conspiracy ineffective side-effect & conspiracy & side-effect \\ 
\hline
SemEval & I used to make the peanut butter energy balls all the time. My famjam loved them! & joy, love & joy, love & joy, love, optimism \\ 
\hline
\end{tabularx}
\caption{Examples showing errors in predictions by our model and the closest baseline (Renee).}

\label{tbl:error-analysis}
\end{table}




Sometime due to limitations in input context, our model may not predict all corresponding labels accurately. Instead, it tends to predict a subset of labels.
Our model outperforms the best baseline by accurately distinguishing correlated labels, such as `joy', `love', and `optimism', which frequently co-occur. While the baseline model misclassified  approximately 33\% of samples predicting as, `joy love optimism',  our model correctly predicts for all such samples. One such example is shown in last row of Table \ref{tbl:error-analysis}.

\section{Conclusion}

In this work, we propose a parameter-efficient generative approach equipped with a dual loss objective to tackle the challenging problem of multilabel classification. Our method introduces a novel and domain-agnostic framework that is flexible enough to be adapted across various applications. By leveraging both generative modeling and discriminative supervision, the approach effectively captures label correlations and enhances prediction robustness. Through extensive experiments, we compare our method against several state-of-the-art models and strong baseline systems specifically designed for the task. Our results show that \modelname~achieves significant performance gains, demonstrating its superiority across multiple evaluation metrics. Furthermore, we conduct detailed ablation studies and empirical analyses to validate the contribution of each component within the framework.

 \section{Limitations}
A limitation of our proposed framework is that the approach relies on the availability of label descriptions, which may not always be readily accessible and would need to be generated when absent. Additionally  it has not been tested on extreme multi-label classification datasets. 

\bibliography{custom}



\end{document}